\ifdefined\pdfobjcompresslevel\pdfobjcompresslevel=0\fi

\documentclass[sigconf]{acmart}

\AtBeginDocument{%
  }

\usepackage{amsmath}

\usepackage{graphicx} 
\usepackage{enumitem}
\usepackage{algorithm}
\usepackage{amsfonts}
\usepackage{color}
\usepackage{algorithmicx}
\usepackage{algpseudocode}
\usepackage{amsmath}

\usepackage{makecell}
\usepackage{float} 
\usepackage{pifont}

\usepackage{placeins} 

\usepackage{tabularx} 

\usepackage{booktabs} 

\usepackage{diagbox}

\usepackage{caption} 

\usepackage{svg}

\usepackage{multirow} 

\usepackage{subcaption}

\usepackage{xcolor}
\newcommand{\highlightgrey}[1]{\colorbox{gray!30}{#1}}

\newcommand{\Indent}{\hspace{1.0em}}

\newcommand\eat[1]{}

\copyrightyear{2026}
\acmYear{2026}
\setcopyright{cc}
\setcctype{by}
\acmConference[CIKM '26]{Proceedings of the 35th ACM International Conference on Information and Knowledge Management}{November 07--11, 2026}{Rome, Italy}
\acmBooktitle{Proceedings of the 35th ACM International Conference on Information and Knowledge Management (CIKM '26), November 07--11, 2026, Rome, Italy}
\acmDOI{10.1145/3799682.3840785}
\acmISBN{979-8-4007-2539-5/2026/11}

\usepackage{amsmath,amsfonts,bm}

\def\eqref#1{equation~\ref{#1}}

\def\1{\bm{1}}

\def\vg{{\bm{g}}}

\def\vw{{\bm{w}}}

\def\vDelta{{\bm{\Delta}}}

\def\mW{{\bm{W}}}

\DeclareMathAlphabet{\mathsfit}{\encodingdefault}{\sfdefault}{m}{sl}
\SetMathAlphabet{\mathsfit}{bold}{\encodingdefault}{\sfdefault}{bx}{n}
\newcommand{\tens}[1]{\bm{\mathsfit{#1}}}
\def\tA{{\tens{A}}}

\def\sR{{\mathbb{R}}}

\newcommand{\Var}{\mathrm{Var}}

\begin{document}

\title{\textsc{Fda-Opt}: Federated Fine-Tuning via Dynamic Update Schedules}


\author{Michael Theologitis}
\authornote{Corresponding author.}
\affiliation{%
 \institution{University of Washington}
 \city{Seattle}
 \state{WA}
 \country{USA}}
\email{mthe@cs.washington.edu}

\author{Vasilis Samoladas}
\affiliation{%
 \institution{Technical University of Crete}
 \city{Chania}
 \country{Greece}}
\email{vsamoladas@tuc.gr}

\author{Antonios Deligiannakis}
\affiliation{%
 \institution{Technical University of Crete}
 \city{Chania}
 \country{Greece}}
\email{adeligiannakis@tuc.gr}

\renewcommand{\shortauthors}{Michael Theologitis, Vasilis Samoladas, \& Antonios Deligiannakis}

\begin{abstract}
  Federated Learning (FL) enables the utilization of vast, previously inaccessible data sources. At the same time, pre-trained Language Models (LMs) have taken the world by storm and for good reason. They exhibit remarkable emergent abilities and are readily adapted to downstream tasks. This opens one of the most exciting frontiers in FL: fine-tuning LMs. Yet, a persistent challenge in FL is the frequent, rigid communication of parameters---a problem magnified by the sheer size of these contemporary models. The \mbox{\textsc{FedOpt}} family of algorithms has become the go-to approach for FL, relying on fixed but arbitrary intervals for model exchanges. Recently, the \textsc{Fda} algorithm prescribed a dynamic approach by monitoring the training progress. However, it introduced a hard-to-calibrate parameter and imposed a rigid synchronization scheme. In this work, we address these limitations by proposing the \mbox{\textsc{Fda-Opt}} family of algorithms---a unified generalization of both \textsc{Fda} and \mbox{\textsc{FedOpt}}. Our experimental evaluation focuses on fine-tuning LMs on downstream NLP tasks and demonstrates that \mbox{\textsc{Fda-Opt}} outperforms \mbox{\textsc{FedOpt}} even when it is configured with hyper-parameters specifically optimized for the latter. In other words, we show that \mbox{\textsc{Fda-Opt}} is a practical, drop-in replacement for \mbox{\textsc{FedOpt}} in modern FL libraries and systems: it requires no additional configuration and delivers superior performance out of the box.
\end{abstract}


\definecolor{darkgreen}{RGB}{0,100,0}


\maketitle

\section{Introduction}

In today's data-driven landscape, leveraging data effectively has become both an opportunity and a challenge. Sensitive data, such as patient records in healthcare or personal information from mobile applications, is often siloed due to privacy concerns~\cite{yan2024FedRoLA, jiang2024dp} and stringent regulations~\cite{chai2025flregulations}. While necessary for protecting user rights, these restrictions hinder the potential of machine learning solutions, as they prevent data from being centralized for training. Federated Learning (FL)~\cite{li2023flsurvey, mcmahan2017FedAvg} emerges as a transformative paradigm by enabling models to be trained directly on the devices or organizations that own the data, bypassing restrictions and privacy concerns.

At the same time, pre-trained Language Models (LMs) have demonstrated extraordinary generalization power that enables them to be readily adapted---through techniques like fine-tuning---to a broad range of downstream tasks~\cite{xu2025llms, hu2024pretrained}. By following the FL paradigm, these powerful models can be trained on vast, previously untapped data sources, enabling new breakthroughs and applications in natural language processing (NLP)~\cite{woise2024thegoodthebadtheugly, wu2024LLMfinetuningFL}. However, FL introduces a fundamental challenge: the frequent exchange of model updates. This issue is only exacerbated by the immense size of modern transformer-based models, making communication overhead a critical bottleneck~\cite{wois2024surveryFoundationModelsFL}.

\begin{figure}[htbp]
  \centering
  \includegraphics[width=0.47\textwidth]{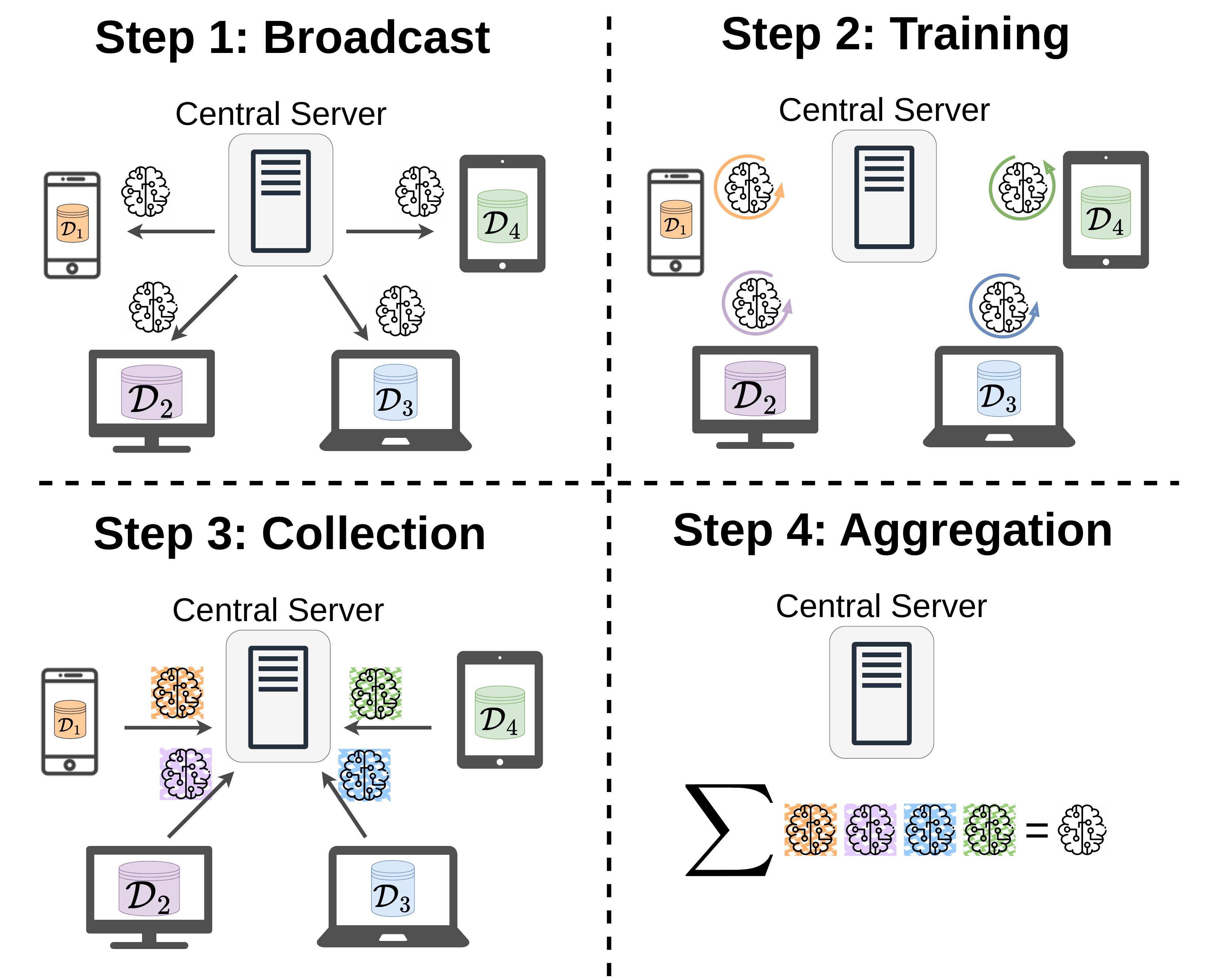}
  \caption{A Federated Learning Round}
  \label{fig:FL_Round}
\end{figure}

To better understand this challenge, it is crucial to examine how FL operates in practice. In a typical FL setting, a central server orchestrates the training process across multiple data owners. Training proceeds in iterative rounds (Figure~\ref{fig:FL_Round}), each consisting of four key steps: broadcast, local training, collection, and aggregation. The server first broadcasts a global model to participating clients, who then train the model locally using their private data. After training, clients send their updates back to the server, who aggregates them into a new global model. These FL training rounds are repeated thousands of times until the model converges. Despite its conceptual simplicity, a critical open question remains: \textit{When should the server collect model updates? When should a round end?}

The widely adopted answer to this question is the \textsc{FedOpt} family of algorithms~\cite{reddi2021fedopt}, which prescribe ending rounds after a fixed number of local steps or epochs. In this approach, each client processes its dataset one or more times before sending model updates back to the server. However, this method has two significant limitations. First, it assumes the training process is static, with configurations predefined throughout training, without taking into account real-time training dynamics. Second, the choice of the number of local steps is arbitrary and lacks justification. Lower values may improve convergence but skyrocket communication overhead, while higher values reduce communication cost but risk unpredictable training behavior or, in many cases, non-convergence \cite{yu2018parallel}. 

 Recently, this trade-off has motivated the development of a dynamic approach to round termination: Federated Dynamic Averaging (\textsc{Fda})~\cite{edbt2025fda}. Unlike static methods, \textsc{Fda} monitors the training state in real-time and adjusts the duration of each round accordingly. Specifically, \textsc{Fda} tracks the \textit{model variance}---a gauge of whether training is progressing well or poorly---and ensures it remains below a prespecified threshold. As long as the variance is within this threshold, local training continues. When the variance exceeds the threshold, the round is terminated, and the server collects the models. This adaptive approach has been shown to significantly reduce communication overhead, achieving orders-of-magnitude improvements in communication compared to \textsc{FedOpt} algorithms.
 
However, the original formulation of \textsc{Fda}~\cite{edbt2025fda} has certain limitations. First, it enforces a rigid synchronization scheme, requiring clients to exchange information with the server synchronously after every learning step, which can bottleneck a real-world FL system. Second, while \textsc{Fda} successfully eliminates the need to tune the number of local steps, it introduces a new parameter---the variance threshold---which is highly nontrivial to calibrate. In addition to these limitations, the experimental comparisons between \textsc{Fda} and \textsc{FedOpt} were between best-case configurations for each algorithm, with different optimizers and hyper-parameters, raising the question whether \textsc{Fda} consistently outperforms \textsc{FedOpt} in an apples-to-apples scenario.

In this work, we propose the \textsc{Fda-Opt} family of algorithms that generalizes both \textsc{Fda} and \textsc{FedOpt} and addresses their core shortcomings. Most importantly, we demonstrate that our proposed \textsc{Fda-Opt} algorithms can be directly plugged into modern FL libraries as drop-in replacements for \textsc{FedOpt}. They require no additional configuration and, in fact, perform better even when using hyper-parameters originally tuned for \textsc{FedOpt}. Our contributions are as follows:
\begin{itemize}[leftmargin=10pt, itemsep=4pt, parsep=0pt]
    \item We propose the \textsc{Fda-Opt} family of algorithms, a unified generalization of both \textsc{Fda} and \textsc{FedOpt}. In doing so, we introduce a dynamic scheme for the variance threshold, removing the need for any manual configuration, and completely alleviate the original synchronization bottleneck.
    
    \item We show that \textsc{Fda-Opt} outperforms \textsc{FedOpt} in communication-efficiency, achieving improvements of at least $2\times$ while operating under conditions optimized for \textsc{FedOpt} (the competitor). Specifically, we manually identify the optimal hyper-parameter configurations for each \textsc{FedOpt} algorithm, and then apply the same configurations to our proposed \textsc{Fda-Opt} counterparts. Remarkably, even with these ``unfair'' and ``rigged'' hyper-paremeters, \textsc{Fda-Opt} achieves $2\times$ greater communication-efficiency for the same model performance. This has the important implication that hyper-parameters proven effective for \textsc{FedOpt} in the literature can be directly leveraged to configure our proposed algorithms.

    \item We show that all \textsc{Fda-Opt} algorithms converge to $5\times$–$10\times$ lower training loss than their \textsc{FedOpt} counterparts within the same number of rounds.
\end{itemize}

The remainder of this paper is organized as follows. Section~\ref{sec:preliminaries} introduces the foundational concepts of Federated Learning (FL), including the \textsc{FedOpt} family of algorithms and the dynamic variance-based \textsc{Fda} approach. Section~\ref{sec:fdaopt} presents our proposed \textsc{Fda-Opt} algorithm. Section~\ref{sec:experimental_approach} outlines the experimental approach. Section~\ref{sec:results_analysis} provides a detailed analysis of experimental results. Section~\ref{sec:related_work} reviews related work. Finally, Section~\ref{sec:conclusion} summarizes the main findings of this work.

\section{Preliminaries}\label{sec:preliminaries}

\vspace{1.5mm}
Before diving into our proposed algorithm, it is essential to understand the key concepts and methods that form its foundation. In this section, we break down the FL training process, explore the widely adopted \textsc{FedOpt} family of algorithms, and outline the recently proposed dynamic \textsc{Fda} approach. Together, these elements set the stage for understanding our subsequent contributions and the idea behind our generalized \textsc{Fda-Opt} algorithm in Section~\ref{sec:fdaopt}.

\subsection{Federated Learning Basics}

\vspace{1.5mm}
Consider a scenario where we aim to train deep neural networks on data distributed across multiple devices or organizations---such as smartphones, hospitals, or sensors---that cannot be exchanged due to constraints like privacy or regulatory restrictions. Federated Learning (FL) addresses this challenge by enabling a set of clients, $\mathcal{K}$, to collaboratively train a deep learning model without sharing their data. Each client $k \in \mathcal{K}$ retains its private local dataset, $\mathcal{D}_k$.

Instead of centralizing data on a powerful cluster for training, FL reverses the traditional approach: training and computation are brought to the data, that is, to the clients. Each client trains a model locally and shares updates with a central server, which consolidates them into a common global model, $\vw \in \mathbb{R}^d$. The collective goal is to minimize the overall training loss across all clients, which can be expressed as the following distributed optimization problem~\cite{wois2024surveryFoundationModelsFL}:

\begin{equation}\label{eq:optimization_prob}
    \underset{\vw \in \sR^d}{\mathrm{minimize}} \; \; F(\vw) \overset{\Delta}{=} \frac{1}{|\mathcal{K}|} \sum_{k \in \mathcal{K}} F_k(\vw)
\end{equation}

\noindent Here, $F_k(\vw) \overset{\Delta}{=} \mathbb{E}_{\zeta_k \sim \mathcal{D}_k} \left[ \ell(\vw; \zeta_k) \right]$ is the local objective function for client $k$, and $\ell(\vw; \zeta_k)$ denotes the loss for a data sample $\zeta_k$ given the model $\vw$.

\subsection{Generalized Federated Averaging}\label{sec:fedopt}

\vspace{1.5mm}
The most widely used algorithm for solving (\ref{eq:optimization_prob}) is Federated Averaging (\textsc{FedAvg})~\cite{mcmahan2017FedAvg}. This method divides the FL training process into iterative rounds. At the beginning of round $t$, the server broadcasts the current global model $\vw_t$ to a selected cohort of participating clients $\mathcal{S}_t \subseteq \mathcal{K}$. Each client $k \in \mathcal{S}_t$ initializes its local model as $\vw^{(k)}_{t, 0} = \vw_t$, and performs $\tau$ local updates of Stochastic Gradient Descent (SGD) using its private dataset $\mathcal{D}_k$. After training, the client sends its updated model $\vw^{(k)}_{t, \tau}$ back to the server. The server then aggregates the updates---typically via averaging\footnote{To be precise, the aggregation is a weighted average based on the number of samples $|\mathcal{D}_k|$ owned by each client; however, to simplify notation, and because the extension to a weighted average is straightforward, we will use averaging throughout this paper.}---to form the new global model, $\vw_{t+1}$.

An indirect yet highly effective way to mitigate the communication burden of FL is to accelerate convergence which led to efforts to incorporate accelerated optimization techniques---such as \textsc{Adam}~\cite{kingma2017adam}---into the FL setting. However, FL clients typically lack enough data for robust statistical coverage and are often stateless, making adaptive optimization at the client level impractical. Instead, the key idea is to shift adaptive optimization to the server.

Specifically, clients still perform $\tau$ local update steps of SGD, but instead of sending back the final model weights themselves, each client computes its local \textit{model change} (or \textit{drift})
\[
\vDelta^{(k)}_{t, \tau} = \vw^{(k)}_{t, \tau} - \vw^{(k)}_{t, 0},
\]
and sends this to the server. The server then averages the updates across clients to obtain the average update direction
\[
\vg_{t, \tau} = -\frac{1}{|\mathcal{S}_t|} \sum_{k \in \mathcal{S}_t} \vDelta^{(k)}_{t, \tau},
\]
which we refer to as the ``pseudo-gradient''. The term ``pseudo'' is used because even though $\vg_{t, \tau}$ is not a traditional gradient---it was not derived from a loss function via backpropagation---it can still be treated in the same way~\cite{reddi2021fedopt}. The server then applies a server-side optimizer, \textsc{ServerOpt}, to update the global model using this ``pseudo''-gradient. Importantly, since $\vg_{t, \tau}$ aggregates updates from multiple clients, it offers sufficient and trustworthy statistics, and the server can easily maintain the optimizer state across rounds. In effect, this addresses the two challenges that make adaptive methods impractical on clients: unreliable statistics, and statelessness.

This approach underpins the \textsc{FedOpt} family of algorithms~\cite{reddi2021fedopt}, which correspond to the non-highlighted lines of Algorithm~\ref{alg:fda-opt} in Section~\ref{sec:fdaopt}, alongside our proposed \textsc{Fda-Opt} algorithm (highlighted lines). We present them together, as \textsc{Fda-Opt} generalizes the core functionality of \textsc{FedOpt}. Additionally, Table~\ref{tab:FedOpt} summarizes the \textsc{FedOpt} family. Notably, \textsc{FedAvg}---which simply averages updates---is mathematically equivalent to SGD with a learning rate of $1.0$.


\begin{table}[t]
\footnotesize
\centering

\caption{Overview of the \textsc{FedOpt} and \textsc{Fda-Opt} families of algorithms. Depending on \textsc{ClientOpt} and \textsc{ServerOpt}, a different name is derived for the FL algorithm. \textsc{Fda-Opt} is our proposed algorithm, introduced in Section~\ref{sec:fdaopt}.}

\label{tab:FedOpt}

\setlength{\tabcolsep}{5pt}
\renewcommand{\arraystretch}{1.0}

\resizebox{\columnwidth}{!}{%
\begin{tabular}{llll}
\toprule

\textbf{\textsc{FedOpt}}
& \textbf{\textsc{Fda-Opt}} (ours)
& \textbf{\textsc{ClientOpt}}
& \textbf{\textsc{ServerOpt}} \\

\midrule

\textsc{FedAvg}~\cite{mcmahan2017FedAvg}
& \textsc{Fda-SGD}
& \textsc{SGD}
& \textsc{SGD} (lr $= 1.0$) \\

\textsc{FedAvgM}~\cite{hsu2019fedAvgM}
& \textsc{Fda-SGDM}
& \textsc{SGD}
& \textsc{SGDM}~\cite{suts2013sgdm} \\

\textsc{FedAdam}~\cite{reddi2021fedopt}
& \textsc{Fda-Adam}
& \textsc{SGD}
& \textsc{Adam}~\cite{kingma2017adam} \\

\textsc{FedAdamW}~\cite{woise2024thegoodthebadtheugly}
& \textsc{Fda-AdamW}
& \textsc{SGD}
& \textsc{AdamW}~\cite{loshchilov2018adamw} \\

\textsc{FedAdaGrad}~\cite{reddi2021fedopt}
& \textsc{Fda-AdaGrad}
& \textsc{SGD}
& \textsc{AdaGrad}~\cite{duchi2011adagrad} \\

\bottomrule
\end{tabular}
}
\end{table}

\subsection{Federated Dynamic Averaging}\label{sec:original_fda}

\begin{figure}[t]
  \centering
  \includegraphics[width=0.47\textwidth]{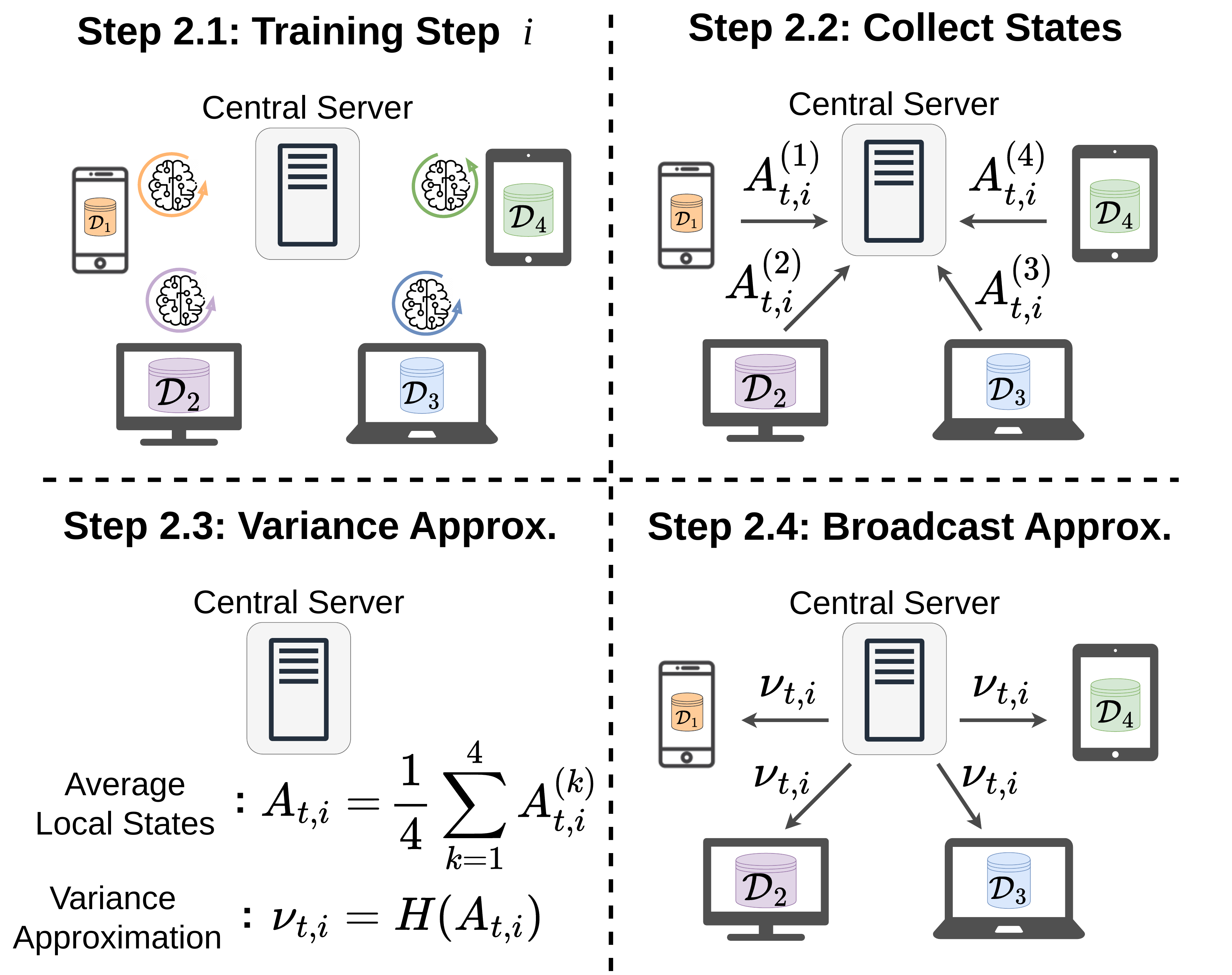}
  \caption{Mechanics of local training, in round $t$, under \textsc{Fda}}
  \label{fig:fda_idea}
\end{figure}

\vspace{1.5mm}
Federated Dynamic Averaging (\textsc{Fda})~\cite{edbt2025fda} is another algorithm for solving (\ref{eq:optimization_prob}), with a key distinction from \textsc{FedAvg}: instead of fixing the number of local steps $\tau$ per round, \textsc{Fda} dynamically decides when clients should stop training and return their models. Specifically, it monitors a metric called \textit{model variance}, and terminates local training once this variance exceeds a predefined threshold $\Theta$.

This paradigm change affects the transition between Step 2 and Step 3 in Figure~\ref{fig:FL_Round}---that is, when clients finish training and the server collects updates. Rather than enforcing a fixed number of SGD steps, \textsc{Fda} allows local training to continue as long as it is stable (i.e., variance remains low), and stops it early when instability is detected.

During each round $t$, clients periodically send small pieces of information to the server, which uses it to estimate the model variance and transmit this estimate back to the clients (see Figure~\ref{fig:fda_idea}). Specifically, after every local step $i$, each client $k \in \mathcal{S}_t$ computes its local model update $\vDelta^{(k)}_{t,i}$ and encodes it into a compact representation:
\begin{align}
    \tA^{(k)}_{t, i} = \left[ \lVert \vDelta^{(k)}_{t, i}  \rVert^2_2 \, , \; \text{sk}(\vDelta^{(k)}_{t, i})  \right] \label{eq:local_state_comp}
\end{align}
where $\text{sk}(\cdot)$ is the AMS sketch operator~\cite{edbt2025fda, cormode2005sketch}. These updates are highly compressed (e.g., in our experiments, a model might be 1GB, but a sketch is just 10KB). The server averages these to form the global state $\tA_{t, i}$, and computes an estimate of the variance $\nu_{t, i}$:
\begin{align*}
    \nu_{t, i} = H(\tA_{t, i}) \quad , \quad \tA_{t, i} = \frac{1}{|\mathcal{S}_t|} \sum_{k \in \mathcal{S}_t} \tA^{(k)}_{t,i}
\end{align*}

Here, $H$ is a function that maps the global state to a variance approximation (we refer to the original work for the full details~\cite{edbt2025fda}). Importantly, if a threshold violation is detected, $\nu_{t, i} > \Theta$, training stops, and the server collects models; otherwise, local training proceeds to the next SGD step, $i+1$, and Steps 2.1--2.4 repeat (Figure~\ref{fig:fda_idea}).

In many ways, \textsc{Fda} resembles the original formulation of \textsc{FedAvg}. In the next section, we will extend its capabilities in a manner analogous to how \textsc{FedOpt} generalized \textsc{FedAvg}.

\section{\textsc{Fda-Opt}: Generalized Federated Dynamic Averaging}\label{sec:fdaopt}

\vspace{1.5mm}
In this section, we propose \textsc{Fda-Opt}, the unified generalization of both \textsc{Fda} and \textsc{FedOpt}, with the following advancements:
\begin{enumerate}[leftmargin=15pt, itemsep=4pt, parsep=0pt]
    \item \textbf{FL Adaptations}: We adapt the algorithm to accommodate FL-specific considerations, including client selection strategies and notation overlooked in the original design.

    \item \textbf{Generalized Averaging}: We enhance server-side aggregation by incorporating adaptive and accelerated optimizers, extending beyond the original simple averaging scheme.

    \item \textbf{Dynamic Variance Threshold}: We introduce a novel dynamic mechanism that automatically calibrates the variance threshold during training, eliminating  the need for any manual tuning.

    \item \textbf{Unified Configuration}: We ensure that \textsc{Fda-Opt} shares the exact same hyper-parameters as \textsc{FedOpt}, allowing it to be configured using well-established settings from prior work.
    
    \item \textbf{Alleviate Synchronization Bottleneck}: We relax the original rigid local-state synchronization requirement by allowing larger intervals between variance approximations.
\end{enumerate}

\subsection{Optimizers}

\vspace{1.5mm}
In the original \textsc{Fda} algorithm, the server-side aggregation relied on a simple averaging scheme. However, extensive empirical evidence highlights the effectiveness of adaptive optimizers in FL~\cite{reddi2021fedopt}. Naturally, we extend the server-side aggregation step by introducing the use of arbitrary optimizers in \textsc{Fda-Opt}, denoted as \textsc{ServerOpt}.

Intuitively, the client optimizer, \textsc{ClientOpt}, focuses on minimizing the local objective for each client based on its private data, while the server optimizer, \textsc{ServerOpt}, operates from a global perspective. By employing adaptive or accelerated optimizers at the server, we can now take advantage of meaningful and accurate statistics, which are inherently unavailable at the client level. 

\subsection{Model Variance}\label{sec:model_variance}

\vspace{1.5mm}
In this subsection, we formalize the notion of model variance, provide intuition for its components, examine how it evolves during training, and explain how it can be interpreted and used as a signal of meaningful progress or instability.

\vspace{2.5mm}
\noindent \textbf{Helpful Notation.} Consider a training round at time $t$, involving a sampled subset of participating clients $\mathcal{S}_t = \{k_1, k_2, \dots, k_N\}$, where $N = |\mathcal{S}_t|$. Moreover, let $i$ be the local training step index. We can organize the client models into a matrix, $\mW_{t, i} \in \sR^{d \times N}$, defined as:

\begin{equation*}
    \mW_{t, i} \overset{\Delta}{=} \left[\vw_{t, i}^{(k_1)}, \vw_{t, i}^{(k_2)}, \ldots , \vw_{t, i}^{(k_N)}\right]
\end{equation*}

\noindent This notation allows us to represent the collective state of the sampled clients in a compact and intuitive manner.

Additionally, as detailed in Section~\ref{sec:fedopt}, each client's model change, $\vDelta^{(k)}_{t, i}$, captures the drift after the $i$-th local training step. Furthermore, their average, $\vg_{t, i}$, can be interpreted as a ``pseudo''-gradient:
\begin{equation*}
    \vg_{t, i} = - \frac{1}{|\mathcal{S}_t|} \sum_{k \in \mathcal{S}_t} \vDelta^{(k)}_{t, i} \quad , \quad \vDelta^{(k)}_{t, i} = \vw^{(k)}_{t, i} - \vw^{(k)}_{t, 0}
\end{equation*}

\vspace{2.5mm}
\noindent \textbf{Definition.} The \textit{model variance} quantifies the spread (or dispersion) of the client models around their average---indicating how compactly they are clustered in parameter space. At round $t$, after $i$ local training steps, it can be written as \cite{edbt2025fda}:
\begin{equation}\label{eq:variance_def}
    \Var(\mW_{t,i}) = \overbrace{\frac{1}{|\mathcal{S}_t|} \sum_{k \in \mathcal{S}_t} \lVert \vDelta^{(k)}_{t, i} \rVert^2_2}^{\text{1\textsuperscript{st} term}} - \underbrace{\left\lVert \vg_{t, i} \right\rVert_2^2}_{\text{2\textsuperscript{nd} term}}
\end{equation}

\vspace{2.5mm}
\noindent \textbf{Intuition.} During a training round $t$, each client $k$ updates its model, resulting in a local change $\vDelta^{(k)}_{t, i}$. The squared norm of this change, $||\vDelta^{(k)}_{t, i}||_2^2$, quantifies how far the client's model has moved from the start of the round (i.e., from the original global model). Essentially, this reflects the amount of information the client has learned locally. The first term in Equation~(\ref{eq:variance_def}), the average of these quantities across all participating clients, represents the \textit{collective amount of information learned}.

However, a large first term does not necessarily mean that training is progressing well, as clients may be moving in conflicting directions, canceling out their updates by averaging. What ultimately matters is the global change $\vg_{t,i}$ obtained after averaging---it foreshadows the information that will be retained in the global model. This is captured by the second term, $||\vg_{t,i}||_2^2$, which inherently accounts for the \textit{direction} of the local changes.

The variance, as defined in Equation~(\ref{eq:variance_def}), reflects the interplay between these two quantities serves as an insightful gauge of the state of the training progress:
\begin{enumerate}[leftmargin=15pt, itemsep=4pt, parsep=0pt]
    \item \textbf{Low Variance}\eat{: It is \textit{harmless} since it} indicates either minimal local progress (both terms low), or substantial local progress in a cohesive and promising way (both terms high).
    \item \textbf{High Variance}\eat{: It is \textit{harmful} as it} indicates significant local progress (high first term) but towards conflicting local minima (low second term).
\end{enumerate}

\begin{figure}[htbp]
    \centering
    \begin{subfigure}[t]{0.49\textwidth}
        \centering
        \includegraphics[width=\textwidth]{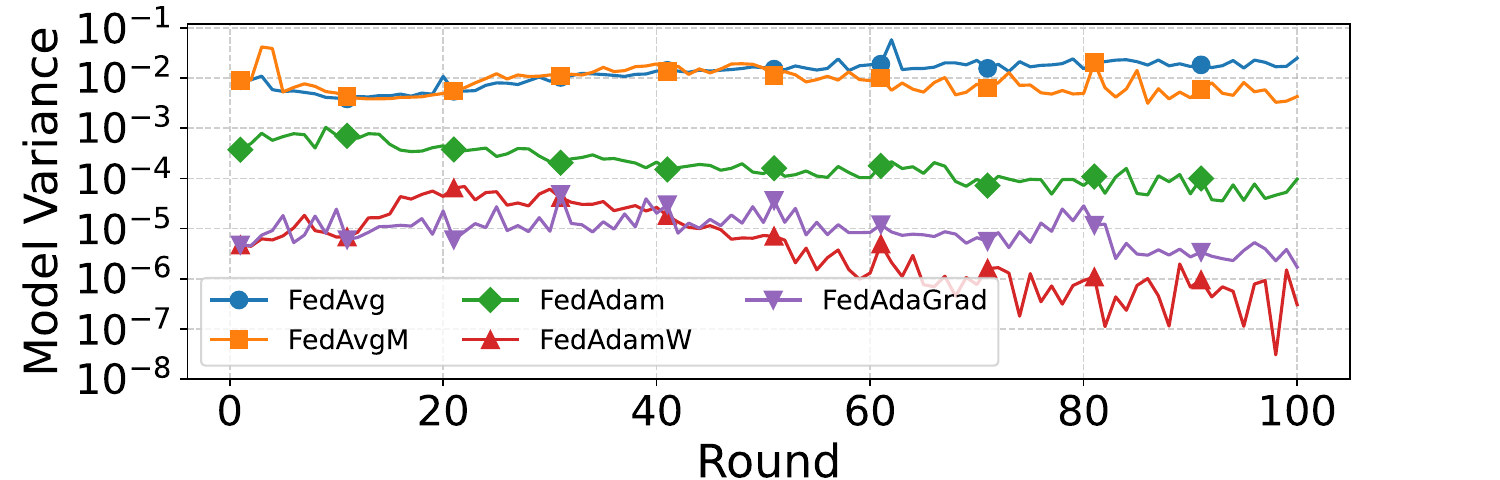}
    \end{subfigure}
    \caption{Variance progression during the first $100$ rounds of training with \textsc{FedOpt} using RoBERTa on the MRPC dataset}
    \label{fig:variance_trends}
\end{figure}

\vspace{2.5mm}
\noindent \textbf{Trends.} What do we really mean by \textit{low} or \textit{high} variance? The answer is that these terms are inherently \textit{relative} and task-specific, influenced by factors like the dataset, model size, optimizer, and hyper-parameters. To illustrate this, we investigate the value of the variance at the end of each FL round under the fixed-round termination schedule of \textsc{FedOpt}. Figure~\ref{fig:variance_trends} plots the variance at the exact moment when model updates are collected from the server. While the experimental setup---including the specific model and datasets---will be detailed later, we present this plot here to help readers intuitively grasp the variance as a metric and its trends.

\eat{To illustrate this, Figure~\ref{fig:variance_trends} plots the variance at the end of each FL training round for a specific model and dataset using different \textsc{FedOpt} algorithms. While the experimental setup will be detailed later, we include this plot here to help readers intuitively grasp the variance as a metric and its trends.}

This plot underscores the critical need to revisit the originally proposed static variance monitoring scheme. Variance trends are inherently dynamic, evolving throughout training, and thus require an adaptive approach to be captured effectively. Moreover, different algorithms exhibit vastly different variance magnitudes (please note that the y-axis is logarithmic) and behaviors. Hence, variance trends should only be interpreted within the context of the same training run. Additionally, the magnitude of the variance may evolve even for the same optimizer by 1-3 orders of magnitude, depending on whether the training is in its early, middle, or late stages.

All in all, \textit{high} variance does not necessarily indicate that the model will converge to a suboptimal solution or that training is progressing poorly. Rather, high variance is interpreted as a sign of misalignment and instability when viewed \textit{in relation} to earlier training rounds of the same run/optimizer.

\subsection{The \textsc{Fda-Opt} Algorithm}

\vspace{1.5mm}
\noindent \textbf{High-Level Overview.} Having established model variance as a key indicator of training dynamics---with distinct behaviors and trends---we now leverage these insights for a unique goal: \emph{to extend local training duration beyond traditional limits without compromising convergence}. To this end, we propose \textsc{Fda-Opt}, presented in Algorithm~\ref{alg:fda-opt}. The highlighted lines represent the core additions that distinguish \textsc{Fda-Opt}: namely, the logic for monitoring model variance and dynamically terminating rounds. These additions act as an adapter to the original \textsc{FedOpt} framework (non-highlighted lines, also detailed in Section~\ref{sec:fedopt}), effectively generalizing it with an dynamic round termination scheme.

\vspace{2.5mm}
\noindent \textbf{Operational Details.} The algorithm begins with a user-provided local training duration $\tau$.  This value is expected to be based on well-established and empirically effective configurations from prior \textsc{FedOpt} work (e.g., \cite{lin2022fedNLP}). Then, \textsc{Fda-Opt} modifies this internally by extending it to a larger value $\tilde{\tau} \gg \tau$ (e.g., $\tilde{\tau} = 10\cdot \tau$; see Line 2).

From this point onward, the algorithm proceeds using the standard round-based approach described in Section~\ref{sec:fedopt}. At the beginning of each round $t$, a cohort of clients $\mathcal{S}_t$ is sampled. These clients train locally for up to $\tilde{\tau}$ steps (Lines 3–9). At the end of training---whose actual length may vary---we compute the ``pseudo''-gradient from model changes and update the global model accordingly (Lines 18 and 21). This process is repeated for a total of $T$ rounds. What sets \mbox{\textsc{Fda-Opt}} apart is its variance monitoring logic.

Specifically, during local training, clients periodically synchronize compact state information (Lines 11–12) to approximate the model variance (Line 13). However, querying variance too frequently (e.g., after every local step) can completely bottleneck a real-world FL system. This is not due to communication volume (the local state states are small), but due to the synchronous nature of the operation. To mitigate this, we introduce a configurable set of step indices, $\mathcal{I}_\text{query} \subseteq [1, 2, ..., \tilde{\tau}]$, which determines the exact points where variance queries are performed (Line 10). In contrast to the original \textsc{Fda} formulation~\cite{edbt2025fda}, which used $\mathcal{I}_\text{query} = [1, 2, ..., \tilde{\tau}]$ (i.e., checking after every step), our method leaves this set user-defined. In practice, we encourage it to be significantly sparser. As we will show later, in our setup the variance is queried only once per epoch.

Lastly, following the discussion on variance behavior in Section~\ref{sec:model_variance}, we allow the threshold value, $\Theta_t$, to evolve across training rounds. Rather than using a fixed value, the threshold is updated dynamically based on the observed training dynamics. Specifically, we define a function, \textsc{ThresholdAdjust}, which takes as input the current threshold $\Theta_t$, the actual variance value at the end of the round, and the final step index $s_t$ where the round terminated (Lines 19–20). Intuitively, these inputs capture the key information needed to answer two questions: (1) \emph{was the current threshold too high or too low?} and (2) \emph{did the round terminate at a reasonable point (e.g., $s_t = 1$ should raise concerns)?} In the next section, we will explore practical choices for this function. Notably, computing the variance in Line 19 requires no additional work, as it reuses quantities---the drifts and ``pseudo''-gradient---that are already available (Lines 17--18).

\begin{algorithm}[t]
\caption{\highlightgrey{\textsc{Fda-Opt} }}
\label{alg:fda-opt}

\begin{algorithmic}[1] 

\Statex \textbf{Input}: \parbox[t]{\dimexpr\linewidth-3em}{Initial $\vw_0$; $\textsc{ClientOpt}$, $\textsc{ServerOpt}$; Total rounds $T$; Local training steps $\tau$, specifically tuned for \textsc{FedOpt}~\cite{reddi2021fedopt}}
\Statex

\State \hspace{-1mm}\highlightgrey{Set $\Theta_0 = - \, \infty$}

\State \hspace{-1mm}\highlightgrey{Choose extended $\tilde{\tau} \gg \tau$} \Comment{\textsc{Fda-Opt} extends local training}

\State \textbf{for each} round $t = 0, \dots, T-1$ \textbf{do}

\State \Indent Sample a subset $\mathcal{S}_t \subseteq \mathcal{K}$ of clients 

\State \Indent Set $\vw^{(k)}_{t, 0} = \vw_t$ for all $k \in \mathcal{S}_t$

\State \Indent \textbf{for each} client $k \in \mathcal{S}_t$ \textbf{in parallel do}

\State \Indent \Indent \textbf{for each} local step $i = 1, ..., $ \hspace{-0.5mm}\highlightgrey{$\tilde{\tau}$} \textbf{do}

\State \Indent \Indent \Indent Compute gradient estimate $\vg^{(k)}_{t, i-1}$ of $\nabla F_k(\vw^{(k)}_{t, i-1})$

\State \Indent \Indent \Indent  $\vw^{(k)}_{t, i} = \textsc{ClientOpt}(\vw^{(k)}_{t, i-1}, \vg^{(k)}_{t, i-1})$

\State \Indent \Indent \Indent \hspace{-1mm}\highlightgrey{\textbf{If} $i \in \mathcal{I}_\text{query}$ \textbf{then}} \Comment{Query variance now}

\State \Indent \Indent \Indent \hspace{-1mm}\highlightgrey{\Indent Construct local state $\tA^{(k)}_{t, i}$ using (\ref{eq:local_state_comp})}

\State \Indent \Indent \Indent \hspace{-1mm}\highlightgrey{\Indent Aggregate states $\tA_{t, i} =  \frac{1}{|\mathcal{S}_t|} \sum_{k \in \mathcal{S}_t} \tA^{(k)}_{t, i}$}

\State \Indent \Indent \Indent \hspace{-1mm}\highlightgrey{\Indent Variance approx. $\nu_{t, i} = H(\tA_{t, i})$} \Comment{See~\cite{edbt2025fda}}

\State \Indent \Indent \Indent \hspace{-1mm}\highlightgrey{\Indent \textbf{If} $\nu_{t, i} > \Theta_t$ \textbf{then}} \Comment{Threshold violation}

\State \Indent \Indent \Indent \hspace{-1mm}\highlightgrey{\Indent \Indent $s_t = i$}  \Comment{Round's final step index}

\State \Indent \Indent \Indent \hspace{-1mm}\highlightgrey{\Indent \Indent \textbf{break}} \Comment{Training round terminates}

\State \Indent \Indent $\vDelta^{(k)}_{t, s_t} = \vw^{(k)}_{t, s_t} - \vw^{(k)}_{t, 0}$ \Comment{Local model change}

\State \Indent $\vg_{t, s_t} = - \frac{1}{|\mathcal{S}_t|} \sum_{k \in \mathcal{S}_t} \vDelta^{(k)}_{t, s_t}$ \Comment{``Pseudo''-gradient}

\State \Indent \hspace{-1mm}\highlightgrey{$\nu ar_{t} = \Var(\mW_{t, s_t})$} \Comment{Compute actual var. using (\ref{eq:variance_def})}

\State \Indent \hspace{-1mm}\highlightgrey{$\Theta_{t+1} = \textsc{ThresholdAdjust}(\nu ar_{t}, \Theta_{t}, s_t)$}

\State \Indent $\vw_{t+1} = \textsc{ServerOpt}(\vw_t, \vg_{t, s_t})$

\State \Return $\vw_{T}$

\end{algorithmic}

\end{algorithm}

\subsection{Default Configuration for \textsc{Fda-Opt}}

\vspace{1.5mm}
We adopt a conservative approach in the configurations below. These are the exact settings used in our experiments, as they yield stable and reliable performance across all tasks. While we believe there is room for improvement---potentially through more aggressive configurations (e.g., a larger linear coefficient of increase of $\tau$)---we leave such tuning as a direction for future work.

\vspace{2.5mm}
\noindent \textbf{Local Training Extension.} We set $\tilde{\tau}$ to be twice the original $\tau$ plus a large constant. This constant is chosen to be eight times the size of a typical client dataset---denoted by $e$. Formally:
\begin{align}
    \tilde{\tau} = 2 \cdot \tau + 8 \cdot \lceil e \rceil, \quad \text{where} \quad e \approx \frac{1}{|\mathcal{K}|} \sum_{k \in \mathcal{K}} |\mathcal{D}_k| \label{eq:N}
\end{align}

\noindent \textbf{Variance Query Indices.} Within each round, we query the variance once per epoch. This completely alleviates the synchronization bottleneck encountered in the original work~\cite{edbt2025fda}. We define:
\[
\mathcal{I}_\text{query} = \left\{e, 2e, ..., \lfloor \tilde{\tau} / e \rfloor \cdot e \right\}
\]

\vspace{2.5mm}
\noindent \textbf{\textsc{ThresholdAdjust}.} In general, variance within a round $t$ tends to increase as local training progresses. We set the threshold $\Theta_t$ such that it is approximately equal to the expected variance around the midpoint of local training (i.e., at step $\big\lfloor \frac{\tilde{\tau}}{2} \big\rfloor$). Assuming linear growth, we can estimate the midpoint variance of the round $t+1$ using a simple proportion, based on known values from the current round $t$---specifically, the actual variance $\nu ar_t$ and termination step $s_t$ computed in Lines 15 and 19 of Algorithm~\ref{alg:fda-opt}, respectively:
\begin{align*}
& \nu ar_{t} \quad \, \, \, \, \xrightarrow{\text{after}} ~s_t ~~~ \, \text{\small local steps}\\
& \nu ar^{(est)}_{t+1} \, \xrightarrow{\text{after}} ~ \frac{\tilde{\tau}}{2} ~~\, \text{\small local steps}
\end{align*}
With this linear prediction, we set the next round's threshold to the estimated variance at the midpoint of local training, $\nu ar^{(est)}_{t+1}$:
\[
\Theta_{t+1} = \frac{\tilde{\tau}/2}{s_t} \cdot \nu ar_{t}
\]

\section{Experimental Approach}\label{sec:experimental_approach}

\vspace{1.5mm}
In this section, we outline the methodology and rationale behind evaluating the proposed \textsc{Fda-Opt} family of algorithms against \textsc{FedOpt}. Specifically, we detail the infrastructure, datasets, data partitioning strategies, and models used in our study.

\subsection{Infrastructure}

\vspace{1.5mm}
We conduct our experiments using the \textit{transformers} library from Hugging Face~\cite{wold2020transformers} with PyTorch \cite{Paszke2019pytorch}. The code is available at GitHub\footnote{\url{https://github.com/michaeltheologitis/FDA-Opt}}. 
All experiments are performed on a local cluster equipped with 2 NVIDIA A10 GPUs and 46 Intel(R) Xeon(R) Silver 4310 CPUs.

\subsection{Tasks \& Datasets}\label{sec:tasks_datasets}

\vspace{1.5mm}
Natural Language Understanding (NLU) tasks are a core component of evaluating machine learning models in NLP. These tasks test a model's ability to understand, reason, and infer relationships between pieces of text. To conduct our experiments, we select six widely used datasets and tasks from the GLUE~\cite{wang2018glue} benchmark: MRPC~\cite{dolan2005mrpc}, SST-2~\cite{socher2013sst2}, RTE~\cite{wang2018glue}, QNLI~\cite{wang2018glue}, MNLI-m and MNLI-mm~\cite{mnlim2018}---each corresponding to a distinct NLU task. Since the GLUE test sets are unpublished, we follow prior studies \cite{zhang2023fedpetuning} and use the validation sets as the new test sets.

\subsection{Data Partitioning}

A critical aspect of FL is the distribution of data across clients, which is characterized by high heterogeneity. Most NLP datasets are designed for centralized machine learning settings. In this subsection, we describe how we partitioned these datasets for FL.

\vspace{2.5mm}
\noindent \textbf{Clients.} The scale of data federation is determined by the number of clients, which generally falls into two categories: \textit{cross-silo} and \textit{cross-device} \cite{jiang2024crosssilocrossdevice}. In cross-silo FL, the number of clients is small---such as hospitals or financial institutions---often allowing all clients to participate in each training round. For this setting, we use the MRPC and RTE datasets. In contrast, cross-device FL involves a much larger pool of clients, such as mobile or IoT devices, where only a fraction of them can participate in each round. For this setting, we use the SST-2, QNLI, and MNLI datasets. Table~\ref{tab:datasets_client_selection} summarizes the client partitioning and sampling strategies across these datasets.


\begin{table}[htbp]
\centering
\caption{Client Partitioning Strategies per Dataset}
\label{tab:datasets_client_selection}

\setlength{\tabcolsep}{5pt}
\renewcommand{\arraystretch}{1.0}

\resizebox{\columnwidth}{!}{%
\begin{tabular}{lcccccc}
\toprule

& \textbf{MRPC}
& \textbf{RTE}
& \textbf{SST-2}
& \textbf{QNLI}
& \textbf{MNLI-m}
& \textbf{MNLI-mm} \\

\midrule

\textbf{\#Clients}
& 10 & 10 & 100 & 250 & 1000 & 1000 \\

\textbf{Sample Size}
& 10 & 5 & 10 & 10 & 10 & 10 \\

\textbf{Cross-Silo}
& \ding{51} & \ding{51} & \ding{55} & \ding{55} & \ding{55} & \ding{55} \\

\textbf{Cross-Device}
& \ding{55} & \ding{55} & \ding{51} & \ding{51} & \ding{51} & \ding{51} \\

\bottomrule
\end{tabular}
}
\end{table}

\vspace{2.5mm}
\noindent \textbf{Label Distribution.} One of the most common heterogeneity scenarios in FL is the presence of non-IID label distributions, where clients possess data with disproportionately distributed labels. For instance, one client may predominantly have samples of label A, while another may mostly contain label B. This phenomenon frequently occurs in real-world applications and is modeled using the Dirichlet distribution~\cite{li2022nonIID,lin2022fedNLP, qin2023statHetero}. The degree of label imbalance is controlled by the concentration parameter $\alpha$. Larger values of $\alpha$ result in more uniform distributions, with $\alpha \to \infty$ producing a perfectly uniform distribution. Conversely, as $\alpha$ approaches zero, the distributions become highly imbalanced, with each client predominantly holding samples from a single label. In our experiments, we set $\alpha = 1.0$, introducing a considerable level of heterogeneity.

\vspace{2.5mm}
\noindent \textbf{Quantity Distribution.} Another common heterogeneity scenario in FL is quantity-based imbalance, where clients possess vastly different amounts of data. We do not explicitly consider this scenario in our experiments. Nevertheless, while we aim to keep quantity distribution balanced, some low degree of quantity imbalance is inevitably introduced due to the label-based non-IID scheme.

\subsection{Models \& Quality Baselines}

\vspace{1.5mm}
In our experiments, we use the pre-trained transformer-based models RoBERTa~\cite{liu2019roberta} and DeBERTaV3~\cite{he2023debertav3improvingdebertausing}. More specifically, we use the \texttt{roberta-base} and \texttt{microsoft/deberta-v3-base} checkpoints available through Hugging Face~\cite{wold2020transformers}, which contain 125 million and 86 million parameters, respectively. Such encoder models are well-suited for FL, as they can be readily fine-tuned for a wide range of practical downstream NLP tasks~\cite{lin2022fedNLP} (e.g., sentiment analysis, textual entailment, etc.). We fine-tune both models with the \textsc{FedOpt} and \textsc{Fda-Opt} algorithms on the datasets outlined in Section~\ref{sec:tasks_datasets}. 

Of course, our ultimate goal in FL is to train accurate and robust models. Thus, as in most of the  FL literature~\cite{reddi2021fedopt}, we evaluate algorithms based on how efficiently they manage to do that. In other words, instead of comparing final accuracy scores or communication alone, we ask: \textit{how many communication rounds does it take to train a model with a specific target quality?}

To define those targets in a consistent and task-agnostic way, we express them as percentages of the best-known centralized performance. Given both models' widespread adoption, their top-reported performance when fine-tuned in centralized settings is well-documented (see Table~\ref{tab:roberta_deberta_top_metrics}). Naturally, due to the adverse conditions of data federation, FL performance is typically lower than centralized baselines. For example, the highest accuracy ever reported for RoBERTa when fine-tuned on MRPC is 90.2\%; then, a target of 90\% means attaining at least $81.18\%$  accuracy ($90\% \cdot 90.2 = 81.18\%$).

    

\begin{table}[htbp]
\centering
\caption{Best-known accuracy scores ($\uparrow$) for fine-tuning each model on each task in an ideal centralized setting. These values serve as target performance baselines.}
\label{tab:roberta_deberta_top_metrics}

\small
\setlength{\tabcolsep}{5pt}
\renewcommand{\arraystretch}{1.0}

\resizebox{\columnwidth}{!}{%
\begin{tabular}{cccccc}

\multicolumn{4}{c}{\textbf{RoBERTa}~\cite{liu2019roberta}}
& \multicolumn{2}{c}{\textbf{DeBERTaV3}~\cite{he2023debertav3improvingdebertausing}} \\

\cmidrule(lr){1-4}
\cmidrule(lr){5-6}

\textbf{MRPC}$\uparrow$
& \textbf{RTE}$\uparrow$
& \textbf{SST-2}$\uparrow$
& \textbf{QNLI}$\uparrow$
& \textbf{MNLI-m}$\uparrow$
& \textbf{MNLI-mm}$\uparrow$ \\

\midrule

$90.2\%$
& $78.7\%$
& $94.8\%$
& $92.8\%$
& $90.6\%$
& $90.7\%$ \\

\bottomrule
\end{tabular}
}
\end{table}

\subsection{Algorithms \& Hyper-Parameters}

\vspace{1.5mm}
We evaluate the most widely used algorithms from the \textsc{FedOpt} family and compare their performance with our proposed \mbox{\textsc{Fda-Opt}} algorithms. To ensure a fair, apples-to-apples comparison, each \mbox{\textsc{FedOpt}} algorithm is paired with its corresponding \textsc{Fda-Opt} variant using identical parameter configurations. For example, we compare \textsc{FedAdamW} with \textsc{Fda-AdamW}, keeping all optimizer settings fixed. The algorithmic pairs we compare are outlined in Table~\ref{tab:FedOpt}.

Our goal is to conclusively demonstrate that \textsc{Fda-Opt} outperforms \textsc{FedOpt} and that hyper-parameters proven effective for \mbox{\textsc{FedOpt}} not only work for \mbox{\textsc{Fda-Opt}} but also yield better results. To show this, we first find the optimal hyper-parameter configuration for each \mbox{\textsc{FedOpt}} algorithm\footnote{For each \textsc{FedOpt} algorithm and dataset/task combination, we perform an exhaustive grid search to identify the best-performing hyper-parameters. For example, to optimize \mbox{\textsc{FedAdam}} training RoBERTa on MRPC, we evaluate a $5 \times 5$ grid of client and server learning rate combinations---25 configurations in total---and select the one achieving the highest accuracy. Across all models, tasks, and optimizers, this process amounts to roughly 700 distinct training runs. While this substantial experimental effort is not emphasized in the main text due to space constraints, it underpins all reported results.}. Then, we apply these to our proposed \textsc{Fda-Opt} counterparts. Given this approach, evidence that \textsc{Fda-Opt} outperforms \textsc{FedOpt} is particularly meaningful.

Lastly, unless otherwise specified, we set the number of local steps $\tau$ equal to one epoch---that is, $\tau = \lceil e \rceil$ from Equation (\ref{eq:N})---as this is the most common and widely accepted choice in the \textsc{FedOpt} literature~\cite{lin2022fedNLP, reddi2021fedopt, hsu2019fedAvgM}. Moreover, we use a batch size of $8$ and train for an exhaustive number of rounds, $T \in \{100, \dots, 1000\}$, ensuring that training continues well beyond likely convergence.

\section{Experimental Results \& Analysis}\label{sec:results_analysis}

\vspace{1.5mm}
In this section, we evaluate the performance of \textsc{Fda-Opt} against \textsc{FedOpt} across a variety of FL tasks, focusing on two key aspects: communication efficiency and convergence behavior, for the same model quality. Notably, our \textsc{Fda-Opt} algorithms operate under configurations optimized for their \textsc{FedOpt} competitors---which makes the following observed improvements all the more noteworthy.

\subsection{Main Findings}

\vspace{1.5mm}
The main findings of our experimental analyses are the following:
\begin{itemize}[leftmargin=10pt, itemsep=4pt, parsep=0pt]
    \item \textbf{Communication-Efficiency.} \textsc{Fda-Opt} demonstrates significant improvements in communication-efficiency. On average, it is $2.15\times$ more efficient in the cross-silo setting and $1.8\times$ in the cross-device setting to train the highest accuracy models.
    \item \textbf{Convergence.} \textsc{Fda-Opt} converges to $5$–$10\times$ lower training loss than \textsc{FedOpt} within the same number of rounds. The cross-silo setting is characterized by a sharp initial drop in loss, which is less pronounced in the cross-device setting.
    \item \textbf{Stability}. \textsc{Fda-Opt} is more robust across different initial local training step values, $\tau$, while \textsc{FedOpt} often fails to converge. Moreover, the communication improvements mirror those observed in the default case.
\end{itemize}

\subsection{Communication-Efficiency}\label{sec:comm_effic}

\begin{table*}[t]
\centering
\caption{Comparison of \textsc{FedOpt} and \textsc{Fda-Opt} in communication-efficiency for training RoBERTa and DeBERTaV3 across different datasets to achieve target metrics. The table reports the number of rounds required to reach $85\%$, $90\%$, $95\%$, and $99\%$ of the top-reported results in the literature (see Table \ref{tab:roberta_deberta_top_metrics})---naturally, the fewer rounds, the better ($\downarrow$). For example, in the bottom-left cell, \textsc{FedAdaGrad} requires $29$ rounds to achieve $90\%$ of the best-reported accuracy for MRPC (i.e., at least $90\% \cdot 90.2 = 81.18\%$), while \textsc{Fda-AdaGrad} achieves the same in $21$ rounds, making it $1.4\times$ more efficient. Similarly, it has a speedup of $1.5\times$ for $95\%$. The ``Average Speedup'' row is computed by averaging the individual \textsc{Fda-Opt} vs. \textsc{FedOpt} speedup values across each column.}
\label{tab:results}

\setlength{\tabcolsep}{1pt}

\begin{tabularx}{\textwidth}{l *{12}{>{\centering\arraybackslash}X}}
& \multicolumn{8}{c}{\textbf{RoBERTa}}
& \multicolumn{4}{c}{\textbf{DeBERTaV3}} \\
\cmidrule(lr){2-9}\cmidrule(lr){10-13}

& \multicolumn{2}{c}{\textbf{MRPC}}
& \multicolumn{2}{c}{\textbf{RTE}}
& \multicolumn{2}{c}{\textbf{SST-2}}
& \multicolumn{2}{c}{\textbf{QNLI}}
& \multicolumn{2}{c}{\textbf{MNLI-m}}
& \multicolumn{2}{c}{\textbf{MNLI-mm}} \\
\cmidrule(lr){2-3}\cmidrule(lr){4-5}\cmidrule(lr){6-7}
\cmidrule(lr){8-9}\cmidrule(lr){10-11}\cmidrule(lr){12-13}

& \textbf{90\%}$\downarrow$ & \textbf{95\%}$\downarrow$
& \textbf{90\%}$\downarrow$ & \textbf{95\%}$\downarrow$
& \textbf{95\%}$\downarrow$ & \textbf{99\%}$\downarrow$
& \textbf{90\%}$\downarrow$ & \textbf{95\%}$\downarrow$
& \textbf{85\%}$\downarrow$ & \textbf{90\%}$\downarrow$
& \textbf{85\%}$\downarrow$ & \textbf{90\%}$\downarrow$ \\
\midrule

\textbf{\textsc{FedAvg}}
& 26 & 30 & 67 & 103 & 8 & 93 & 35 & 71 & 12 & 16 & 12 & 15 \\

\textbf{\textsc{Fda-SGD} \textnormal{(ours)}}
& \textbf{6} & \textbf{10} & \textbf{16} & \textbf{20}
& \textbf{7} & \textbf{43} & \textbf{9} & \textbf{23}
& \textbf{4} & \textbf{8} & \textbf{4} & \textbf{6} \\

\addlinespace[3pt]

\textbf{\textsc{FedAvgM}}
& 21 & 31 & 111 & 182 & 30 & 91 & 38 & 81 & 29 & 35 & 29 & 35 \\

\textbf{\textsc{Fda-SGDM} \textnormal{(ours)}}
& \textbf{5} & \textbf{16} & \textbf{41} & \textbf{126}
& \textbf{12} & \textbf{41} & \textbf{11} & \textbf{44}
& \textbf{12} & \textbf{15} & \textbf{12} & \textbf{15} \\

\addlinespace[3pt]

\textbf{\textsc{FedAdam}}
& 20 & 40 & 118 & \textbf{153} & 12 & 42 & 11 & 22 & 11 & 12 & 11 & 12 \\

\textbf{\textsc{Fda-Adam} \textnormal{(ours)}}
& \textbf{7} & \textbf{11} & \textbf{72} & 188
& \textbf{8} & \textbf{25} & \textbf{6} & \textbf{20}
& \textbf{9} & \textbf{10} & \textbf{9} & \textbf{10} \\

\addlinespace[3pt]

\textbf{\textsc{FedAdamW}}
& 17 & 25 & 118 & \textbf{160} & 12 & 35 & 11 & 22 & 57 & 66 & 57 & 66 \\

\textbf{\textsc{Fda-AdamW} \textnormal{(ours)}}
& \textbf{14} & \textbf{16} & \textbf{71} & 176
& \textbf{8} & \textbf{14} & \textbf{6} & \textbf{20}
& \textbf{43} & \textbf{54} & \textbf{42} & \textbf{52} \\

\addlinespace[3pt]

\textbf{\textsc{FedAdaGrad}}
& 29 & 35 & 20 & 96 & 21 & 67 & 12 & 24 & \textbf{8} & 15 & 8 & \textbf{11} \\

\textbf{\textsc{Fda-AdaGrad} \textnormal{(ours)}}
& \textbf{21} & \textbf{23} & \textbf{18} & \textbf{54}
& \textbf{7} & \textbf{24} & \textbf{6} & \textbf{20}
& 14 & 15 & \textbf{6} & 14 \\

\midrule
\textbf{Average Speedup}
& \textcolor{darkgreen}{$\bm{+2.8\times}$}
& \textcolor{darkgreen}{$\bm{+2.3\times}$}
& \textcolor{darkgreen}{$\bm{+2.3\times}$}
& \textcolor{darkgreen}{$\bm{+2\times}$}
& \textcolor{darkgreen}{$\bm{+1.9\times}$}
& \textcolor{darkgreen}{$\bm{+2.3\times}$}
& \textcolor{darkgreen}{$\bm{+2.6\times}$}
& \textcolor{darkgreen}{$\bm{+1.6\times}$}
& \textcolor{darkgreen}{$\bm{+1.7\times}$}
& \textcolor{darkgreen}{$\bm{+1.6\times}$}
& \textcolor{darkgreen}{$\bm{+1.9\times}$}
& \textcolor{darkgreen}{$\bm{+1.6\times}$} \\
\bottomrule
\end{tabularx}
\end{table*}

\vspace{1.5mm}
The communication overhead incurred by each algorithm is directly analogous to the number of rounds. For \textsc{FedOpt}, communication between clients and the server occurs only at the end of each round. In \mbox{\textsc{Fda-Opt}}, there is additional communication due to the transmission of small sketches. However, the size of these sketches is negligible compared to the overall communication cost of transmitting language models (it is $\times 10^6$ smaller).

As is standard in the literature~\cite{wang2021fieldguide}, we evaluate our algorithms by comparing their performance against predefined target metrics---such as accuracy ($\uparrow$). The primary objective of the algorithms is to train a model that attains the target metric. Once they first do, we assess their communication efficiency by examining the number of FL rounds they required ($\downarrow$). Table~\ref{tab:results} illustrates this evaluation.

\vspace{2.5mm}
\noindent \textbf{Cross-Silo.} The cross-silo setting includes RoBERTa trained on the MRPC and RTE datasets, as outlined in Table~\ref{tab:datasets_client_selection}. 
\begin{itemize}[leftmargin=10pt, itemsep=4pt, parsep=0pt]
    \item \textbf{RoBERTa on MRPC.} As per Table~\ref{tab:results}, \textsc{Fda-Opt} demonstrates superior communication-efficiency, with an average speedup of $2.3\times$-$2.8\times$ compared to \textsc{FedOpt}. It is $2.8\times$ more efficient in reaching $90\%$ of the target metric, and $2.3\times$ for $95\%$.
    \item \textbf{RoBERTa on RTE.} Similar trends are observed for the RTE dataset. \textsc{Fda-Opt} achieves average speedups of $2\times$-$2.3\times$ over \textsc{FedOpt}, with specific gains of $2.3\times$ for $90\%$, and $2\times$ for $95\%$ of the target metrics.
\end{itemize}
When the goal is to train the most accurate model---defined here as achieving $95\%$ of the top-reported metrics---\textsc{Fda-Opt} is, on average, $2.15\times$ more communication-efficient in the cross-silo setting.

\vspace{2.5mm}
\noindent \textbf{Cross-Device.} In the cross-device setting, we analyze performance of training RoBERTa on the SST-2 and QNLI datasets, and DeBERTaV3 on MNLI-m and MNLI-mm, as outlined in Table~\ref{tab:datasets_client_selection}.
\begin{itemize}[leftmargin=10pt, itemsep=4pt, parsep=0pt]
    \item \textbf{RoBERTa on SST-2.} \textsc{Fda-Opt} consistently outperforms \mbox{\textsc{FedOpt}}, achieving an improvement in communication-efficiency of $1.9\times$-$2.3\times$. Specifically, \textsc{Fda-Opt} is $1.9\times$ more efficient in achieving $90\%$ of the target metric, and $2.3\times$ for $99\%$.

    \item \textbf{RoBERTa on QNLI.} Again, \textsc{Fda-Opt} exhibits speedups ranging from $1.6\times$-$2.6\times$. It is $2.6\times$ more efficient at $90\%$, and $1.6\times$ at $95\%$ of the target metric.

    \item \textbf{DeBERTaV3 on MNLI-m.} Similarly, \textsc{Fda-Opt} has improvements of $1.7\times$ and $1.6\times$, corresponding to $85\%$ and $90\%$ of the target metrics, respectively.

    \item \textbf{DeBERTaV3 on MNLI-mm.} Lastly, \textsc{Fda-Opt} is $1.7\times$ and $1.6\times$ more communication-efficient than \textsc{FedOpt}, for attaining $85\%$ and $90\%$ of the target accuracies, respectively.
\end{itemize}
Similarly, in the cross-device setting, \textsc{Fda-Opt} achieves an average improvement in  communication-efficiency of $1.8\times$ when training models to attain the highest possible accuracy targets.

\subsection{Convergence}\label{sec:convergence}

\begin{figure}
    \centering
    \begin{subfigure}[t]{0.23\textwidth} 
        \centering
        \includegraphics[width=\textwidth]{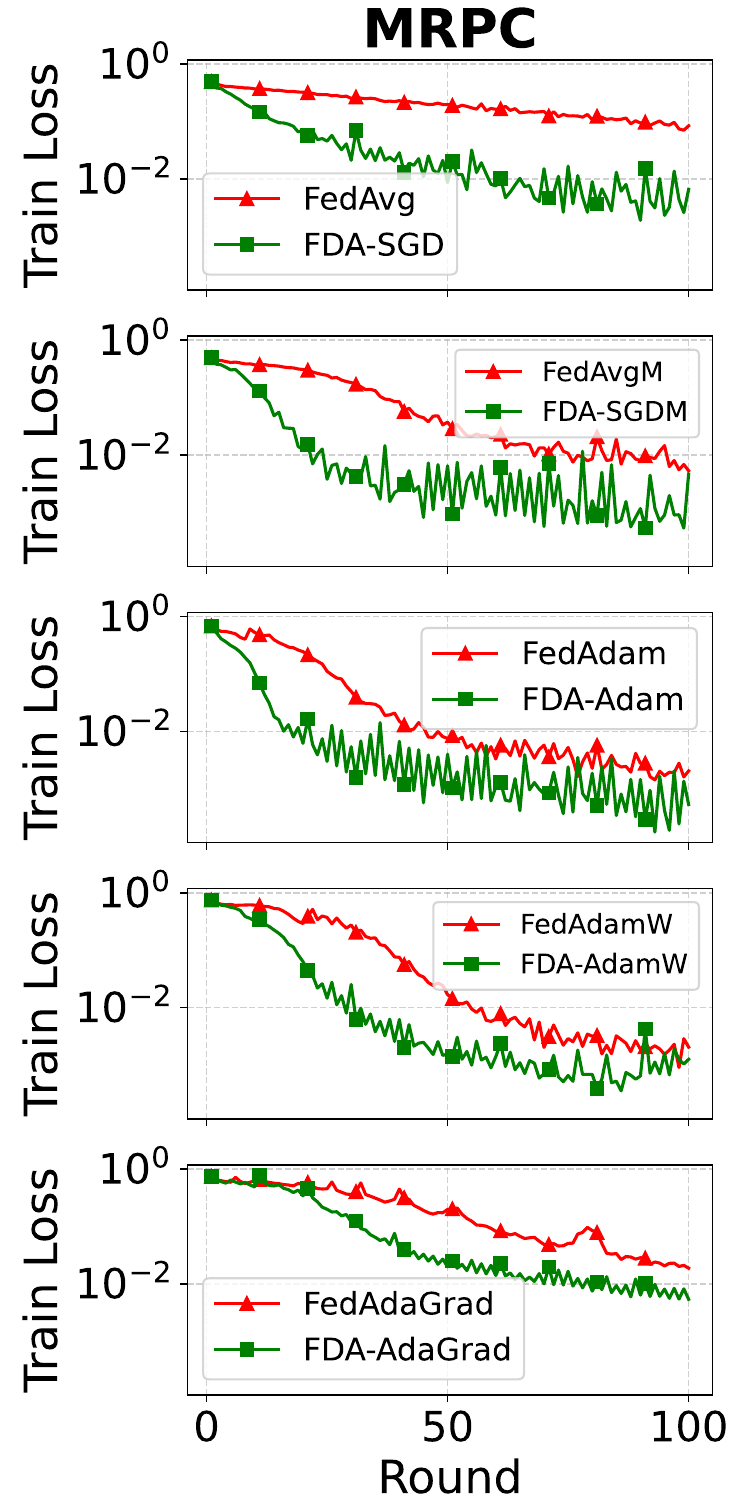}
    \end{subfigure}
    \begin{subfigure}[t]{0.23\textwidth} 
        \centering
        \includegraphics[width=\textwidth]{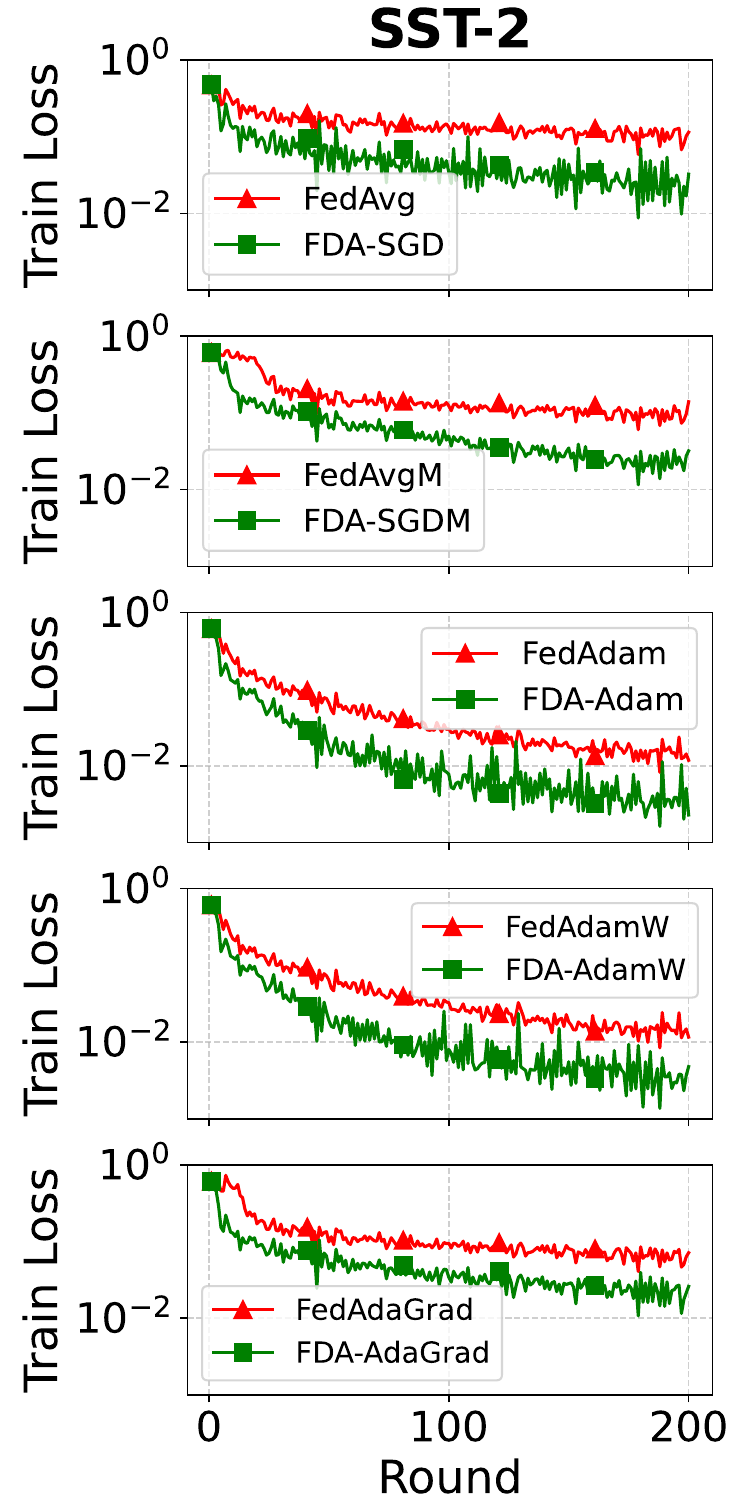}
    \end{subfigure}
    \caption{Training Loss progression of training RoBERTa with \textsc{Fda-Opt} vs. \textsc{FedOpt} on MRPC (left) and SST-2 (right)}
    \label{fig:train_loss_progression}
\end{figure}

\vspace{1.5mm}
Analyzing training loss progression is key to understanding the convergence behavior of FL algorithms. It provides insights into how quickly and effectively models learn---essentially, their ability to minimize objectives. In this subsection we analyze Figure~\ref{fig:train_loss_progression} which compares the performance of \textsc{Fda-Opt} and \textsc{FedOpt} on two representative datasets from cross-silo and cross-device settings. The conclusions drawn regarding convergence pace and quality from these two datasets are indicative of the broader trends observed across all experiments.

\vspace{2.5mm}
\noindent \textbf{Cross-Silo.} On the MRPC dataset---which simulates a cross-silo setting following Table~\ref{tab:datasets_client_selection}---all \textsc{Fda-Opt} algorithms demonstrate significantly faster convergence than their \textsc{FedOpt} counterparts (Figure~\ref{fig:train_loss_progression}). For instance, \textsc{Fda-SGD} achieves a training loss $100\times$ smaller than \textsc{FedAvg} within the first $100$ rounds. Additionally, both \mbox{\textsc{Fda-SGDM}} and \mbox{\textsc{Fda-AdamW}} exhibit sharp drops within the initial $30$–$40$ rounds, outperforming \textsc{FedAvgM} and \textsc{FedAdamW}, respectively, with \mbox{\textsc{Fda-SGDM}} stabilizing at a loss $10\times$ smaller than \textsc{FedAvgM}. Meanwhile, \textsc{Fda-Adam} converges to a loss $5\times$ smaller than \textsc{FedAdam}, and \textsc{Fda-AdaGrad} consistently maintains a $5\times$ lower loss than \mbox{\textsc{FedAdaGrad}} after the initial rounds.

\vspace{2.5mm}
\noindent \textbf{Cross-Device.} On the SST-2 dataset---representing a cross-device setting as outlined in Table~\ref{tab:datasets_client_selection}---all \textsc{Fda-Opt} algorithms consistently achieve $5$–$10\times$ lower training loss compared to their \textsc{FedOpt} counterparts (Figure~\ref{fig:train_loss_progression}). Unlike the cross-silo case, the initial drop in training loss is less pronounced, which is expected since each round involves only a small subset of clients. This leads to initial updates that are less representative of the overall training distribution.

\subsection{Stability to Local Training Extension}

\vspace{1.5mm}
Empirical evidence suggests that increasing the number of local training steps $\tau$ can lead to degraded performance or non-convergence altogether~\cite{wang2021fieldguide, yu2018parallel}. Since our scheme  extends local training even further ($\tilde{\tau} \gg \tau$), it is crucial to examine whether this extension introduces any instability. To this end, we train RoBERTa on MRPC (cross-silo) and SST-2 (cross-device) using a range of values for $\tau$.

\begin{figure}[htbp]
    \centering
    \begin{subfigure}[t]{0.23\textwidth} 
        \centering
        \includegraphics[width=\textwidth]{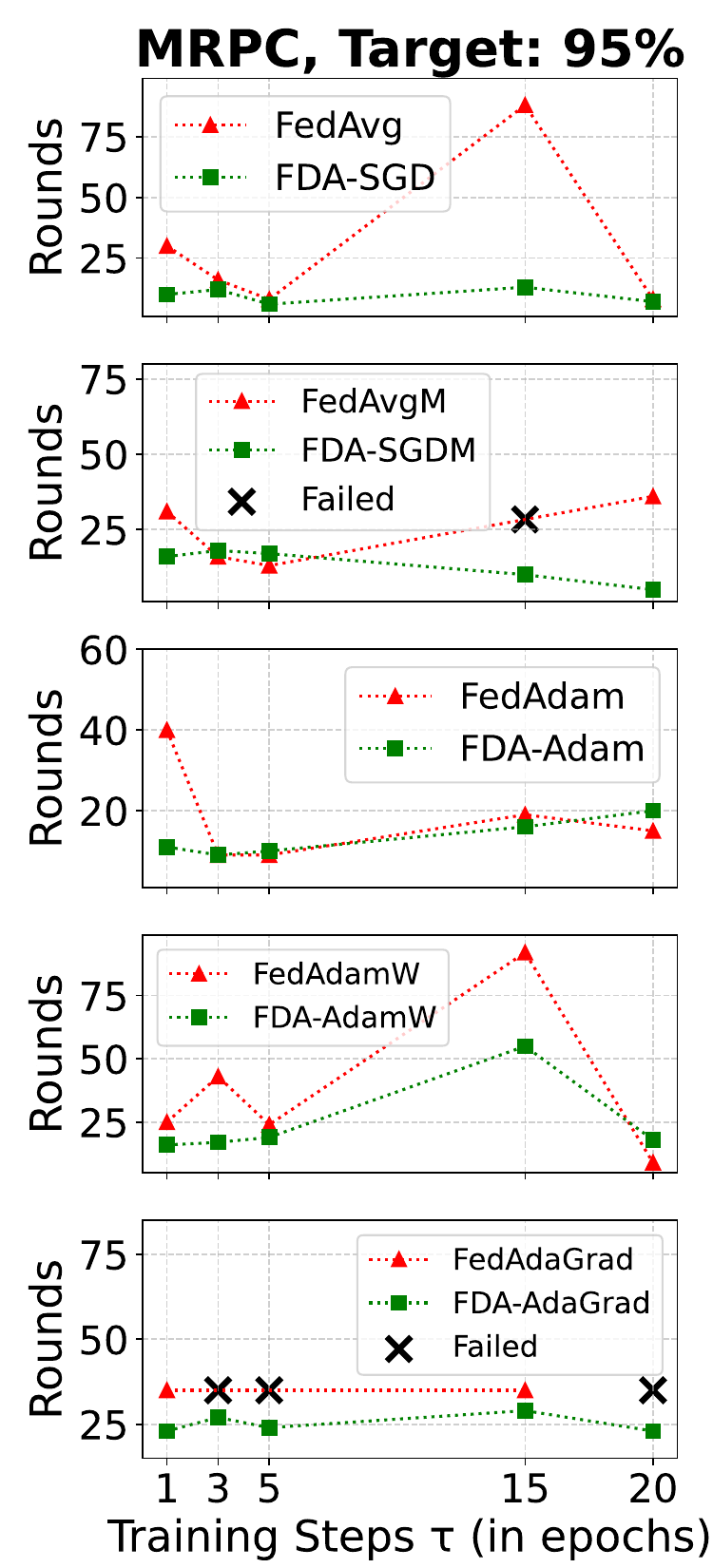}
    \end{subfigure}
    \begin{subfigure}[t]{0.23\textwidth} 
        \centering
        \includegraphics[width=\textwidth]{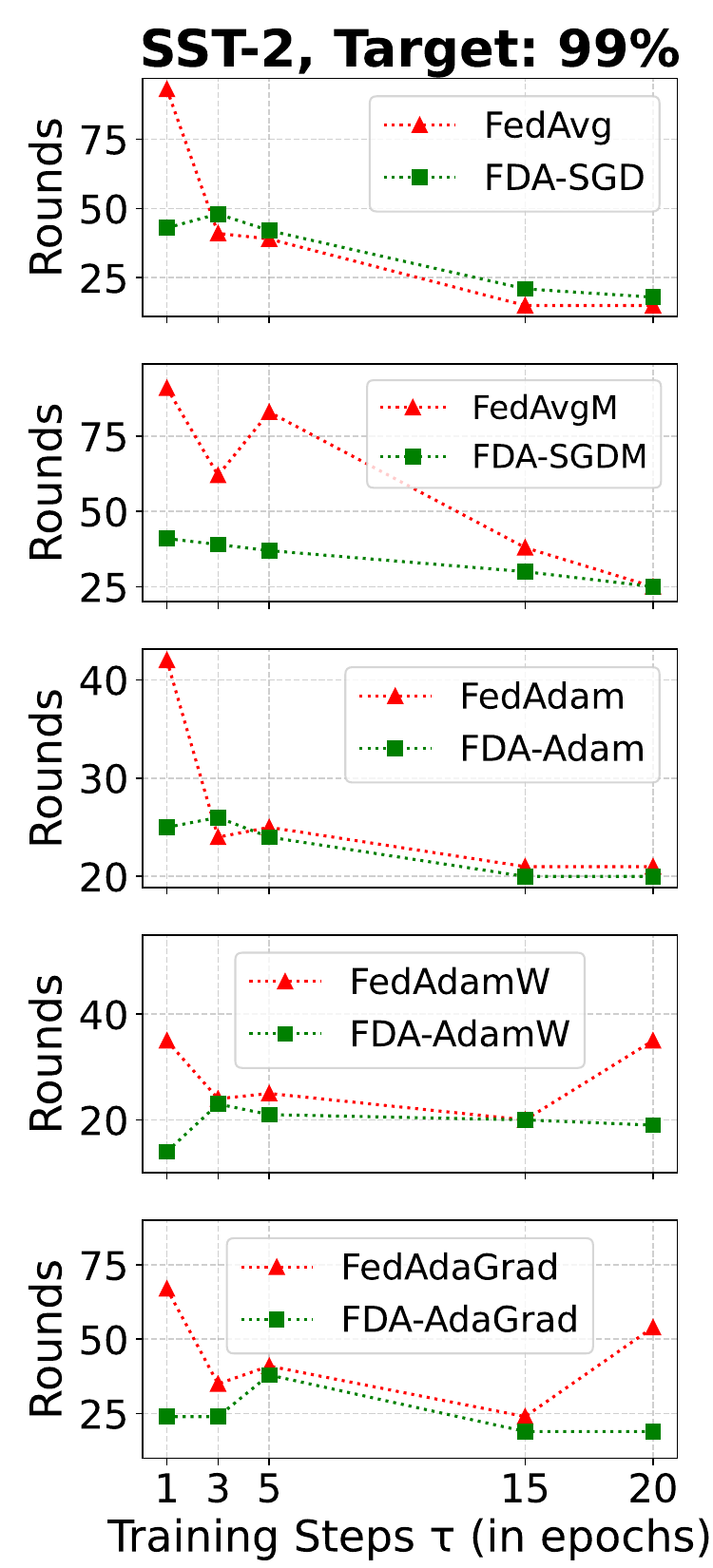}
    \end{subfigure}
    \caption{Training RoBERTa with \textsc{Fda-Opt} vs. \textsc{FedOpt} for varying initial local training steps $\tau$ (measured in epochs). We report the number of rounds required to attain the target accuracy. Failure to converge is marked with ``$\bm{\times}$''}
    \label{fig:varying_E}
\end{figure}

The results, shown in Figure~\ref{fig:varying_E}, reveal two key insights. First, in every one of the 10 subplots, \textsc{Fda-Opt} consistently requires fewer training rounds than \textsc{FedOpt}. Across all 50 individual setups, \textsc{Fda-Opt} outperforms \textsc{FedOpt} in 40 cases---often by a large margin. Conversely, \textsc{FedOpt} shows only marginal improvements in the remaining 10 cases. Second, and most importantly, \textsc{Fda-Opt} converges reliably across all experiments. In contrast, \textsc{FedOpt} fails to converge in 4 out of 50 (specifically, \textsc{FedAvgM} and \textsc{FedAdaGrad} on MRPC). This is particularly problematic in FL, where non-convergence is almost impossible to detect and may lead to severe, and exploding communication cost.

These findings highlight that \textsc{Fda-Opt} is not only more efficient, but also inherently stable and \textit{safe} for extended local training intervals---a direct result of monitoring the variance.

\eat{
\begin{table}[!htbp]
\centering
\caption{Cumulative number of Rounds to attain $95\%$ and $99\%$ of the top-reported accuracy for MRPC and SST-2, respectively, with varying Local Epochs $E$}
\label{tab:total_rounds_E}
\begin{tabular}{|l|l|l|}
\cline{2-3}
\multicolumn{1}{c|}{} & \textbf{MRPC} & \textbf{SST-2} \\ \hline
\textbf{\textsc{FedOpt}} & 601 & 995\\ \hline
\textbf{\textsc{Fda-Opt}} & 431 & 680 \\ \hline \hline

Improvement & $1.5 \times$ & $1.4 \times$ \\ \hline
\end{tabular}
\end{table}
}

\section{Related Work}\label{sec:related_work}

\vspace{1.5mm}
This section reviews related work on communication-efficient FL.

\vspace{2mm}
\noindent \textbf{Round Termination.} Several works have explored strategies for determining the round termination intervals in FL. The study in~\cite{wang2019adaptiveFL} aims to find a fixed, optimal interval by balancing local computation and aggregation constraints. In contrast, \cite{mills2023faster} analyzes the trade-off between fast convergence and round completion rates under strongly convex client objectives, leading to a decaying round duration scheme. Similarly, \cite{wang2018adaptive} minimizes convergence error with respect to wall-clock time by progressively decreasing the round duration. On the other hand, \cite{haddadpour2019localSGD} focuses on minimizing communication rounds for a fixed number of model updates, which results in an increasing sequence of round duration intervals. Unlike these approaches, which rely on predefined schedules---whether fixed, decaying, or increasing---\textsc{Fda-Opt} makes round termination decisions dynamically, in real-time, based on the state of the training.

\vspace{2.5mm}
\noindent \textbf{Convergence.} The most direct way to alleviate the communication burden in FL is to accelerate convergence. This motivated the development of the \textsc{FedOpt} family (Section \ref{sec:fedopt}). However, FL algorithms still struggle with high heterogeneity, where different clients converge to disparate and often conflicting local minima~\cite{kale2019scaffold, reddi2021fedopt, WenigP22}, ultimately slowing down convergence. To address this, \textsc{SCAFFOLD} introduced control variates to correct client drift~\cite{kale2019scaffold}. Similarly, the Mime framework~\cite{karimireddy2021mime} leverages both control variates and server statistics. \textsc{FedProx}\cite{kumar2018fedProx} tackles heterogeneity by adding an $L^2$ regularization term. Notably, it has been demonstrated that \textsc{FedOpt} outperforms \textsc{FedProx} in NLP tasks similar to ours~\cite{lin2022fedNLP}. Lastly, \textsc{FedDyn}~\cite{acar2021fedDyn} introduces a dynamic regularizer, ensuring that client convergence aligns with the stationary point of the global objective. Importantly, \textsc{Fda-Opt} is orthogonal to these optimization methods as we focus on a complementary aspect: when to terminate rounds.

\vspace{2.5mm}
\noindent \textbf{PEFT.} Parameter-efficient fine-tuning (PEFT)~\cite{li2021peft, lester2021peft, hu2022lora, ben-zaken-etal-2022-bitfit} trains only a small subset of a model’s parameters while keeping the rest frozen. One of the most widely used approaches, LoRA~\cite{hu2022lora}, injects low-rank adaptation matrices into a transformer's frozen attention layers, reducing the number of trainable parameters to $< 1\%$. Given the communication constraints in FL, recent studies have explored the applicability of PEFT methods in this setting~\cite{zhang2023fedpetuning, sun2024improvingLoraPrivate}. However, techniques like LoRA struggle with the high heterogeneity present in FL~\cite{sun2024improvingLoraPrivate}. SLoRA~\cite{babakniya2023slora} addresses this issue by leveraging sparse fine-tuning~\cite{ansell-etal-2022-composable} to initialize the injected parameters more effectively. These PEFT methods can be readily integrated with \textsc{Fda-Opt}, as the choice of trainable parameters, $\mathbf{w}$, remains user-defined. 

\vspace{2.5mm}
\noindent \textbf{Compression.} Another way to reduce communication in FL is to compress transmitted information using techniques like quantization~\cite{shlezinger2020quantizationFL} and sparsification~\cite{aji2017sparsification, basu2019qsparselocalsgd}. A comprehensive survey of these methods is provided in~\cite{wang2023compressionInDDL}. Notably, compression is orthogonal to our approach and can be seamlessly integrated with \textsc{Fda-Opt}.

\section{Conclusion}\label{sec:conclusion}

\vspace{1.5mm}
In this work, we introduced the \textsc{Fda-Opt} family of algorithms, a unified generalization of both \textsc{Fda} and \textsc{FedOpt}, addressing their core limitations. We empirically prove that \textsc{Fda-Opt} is more communication efficient and reliable than \textsc{FedOpt}. Furthermore, through carefully designed experiments, we demonstrate that \textsc{Fda-Opt} can seamlessly replace \textsc{FedOpt}, as it can be directly configured using settings from the \textsc{FedOpt} literature. To this end, each comparison between the two algorithmic families was conducted using the best-performing configurations for our competitor, \textsc{FedOpt}---showing that well-established settings from the literature can be applied to our algorithms without modification. Importantly, even under these ``unfair'' conditions, \textsc{Fda-Opt} achieves at least $2\times$ greater communication-efficiency on average and consistently converges to $5\times$–$10\times$ lower training loss. These findings hint at significant practical implications for improving modern FL libraries and real-world deployments.

\section{Acknowledgments}

\vspace{2.5mm}
This research is supported by European Union’s Horizon 2020 Research and Innovation Programme under grant agreement No. 101092749, project ``Critical Action
Planning over Extreme-Scale Data (CREXDATA)''.

\section{GenAI Usage Disclosure}

AI assistants were used solely for polishing and grammar checking.

\bibliographystyle{ACM-Reference-Format}
\bibliography{references}

\end{document}